\documentclass[conference]{IEEEtran}
\IEEEoverridecommandlockouts
\usepackage{cite}
\usepackage{amsmath,amssymb,amsfonts}
\usepackage{algorithmic}
\usepackage{graphicx}
\usepackage{textcomp}
\usepackage{xcolor}
\usepackage{booktabs}
\usepackage{algorithm}
\usepackage{amsfonts}
\usepackage{multirow}
\usepackage{multicol}
\usepackage[table]{xcolor}
\def\BibTeX{{\rm B\kern-.05em{\sc i\kern-.025em b}\kern-.08em
    T\kern-.1667em\lower.7ex\hbox{E}\kern-.125emX}}
\begin{document}

\title{PRISM: A Personality-Driven Multi-Agent Framework for Social Media Simulation}

\author{
    \IEEEauthorblockN{Zhixiang Lu\IEEEauthorrefmark{1}}
    \IEEEauthorblockA{\textit{University of Liverpool}\\
    Liverpool, United Kingdom \\
    zhixiang@liverpool.ac.uk}
    \and
    \IEEEauthorblockN{Xueyuan Deng\IEEEauthorrefmark{1}}
    \IEEEauthorblockA{\textit{University of Texas at Austin}\\
    Texas, United States \\
    xueyuan.deng@utexas.edu}
    \and
    \IEEEauthorblockN{Yiran Liu}
    \IEEEauthorblockA{\textit{University College London}\\
    London, United Kingdom \\
    liuyr1014@outlook.com}
    \and

    \IEEEauthorblockN{Yulong Li}
    \IEEEauthorblockA{\textit{Xi'an Jiaotong-Liverpool University}\\
    Jiangsu, China \\
    yulong.li19@student.xjtlu.edu.cn}
    \and
    \IEEEauthorblockN{Qiang	Yan}
    \IEEEauthorblockA{\textit{Chinese Academy of Sciences}\\ 
    Beijing, China \\
    yanqiang@ict.ac.cn}
    \and
    \IEEEauthorblockN{Imran Razzak}
    \IEEEauthorblockA{\textit{Mohamed bin Zayed University of Artificial Intelligence}\\ 
    Abu Dhabi, United Arab Emirates \\
    imran.razzak@mbzuai.ac.ae}
    \and
    \IEEEauthorblockN{Jionglong Su\IEEEauthorrefmark{2}}
    \IEEEauthorblockA{\textit{Xi'an Jiaotong-Liverpool University}\\
    Jiangsu, China \\
    jionglong.su@xjtlu.edu.cn}
}

    
    

\maketitle
\begingroup
\renewcommand\thefootnote{}
\footnotetext{$^{*}$ Equal contribution, $^{\dagger}$ Corresponding author.}
\endgroup
\begin{abstract}
Traditional agent-based models (ABMs) of opinion dynamics often fail to capture the psychological heterogeneity driving online polarization due to simplistic homogeneity assumptions. This limitation obscures the critical interplay between individual cognitive biases and information propagation, thereby hindering a mechanistic understanding of how ideological divides are amplified. To address this challenge, we introduce the Personality-Refracted Intelligent Simulation Model (PRISM), a hybrid framework coupling stochastic differential equations (SDE) for continuous emotional evolution with a personality-conditional partially observable Markov decision process (PC-POMDP) for discrete decision-making. In contrast to continuous trait approaches, PRISM assigns distinct Myers-Briggs Type Indicator (MBTI) based cognitive policies to multimodal large language model (MLLM) agents, initialized via data-driven priors from large-scale social media datasets. PRISM achieves superior personality consistency aligned with human ground truth, significantly outperforming standard homogeneous and Big Five benchmarks. This framework effectively replicates emergent phenomena such as rational suppression and affective resonance, offering a robust tool for analyzing complex social media ecosystems.

\end{abstract}


\section{Introduction}
\label{sec:introduction}

Social media platforms have evolved into complex adaptive ecosystems where information diffusion and opinion formation are deeply governed by the psychological heterogeneity of their users. However, the modeling of these dynamics is often hindered by the homogeneity assumption inherent in traditional agent-based models (ABMs) \cite{UhrmacherWeyns2018}. Conventional simulations frequently employ agents with uniform interaction rules or simplified behavioristic parameters. Consequently, they fail to capture the psychological heterogeneity that fundamentally shapes online discourse. This limitation becomes particularly critical when modeling discussions on controversial topics. In such contexts, individual differences in cognitive processing, including the tendency to suppress emotion or the propensity for affective contagion, significantly impact polarization dynamics \cite{Chen2019}.

To bridge this gap, the Myers-Briggs Type Indicator (MBTI) \cite{Myers1976} framework offers a robust taxonomy for modeling distinct cognitive architectures. Unlike continuous trait models such as the Big Five Personality Model (BFPM) \cite{mccrae_john_1992} which modulate behavior via scalar parameters, the MBTI categorizes individuals into sixteen distinct types along four dichotomous dimensions including thinking versus feeling \cite{boyle_saklofske_matthews_2015}. This categorical structure is particularly advantageous for computational simulation as it allows for the assignment of discrete and role-based interaction policies. For instance, the system can define distinct behavioral patterns for a debater role associated with the ENTP type versus a mediator role associated with the INFP type \cite{boer_2013}. Recent advances in psycholinguistics have further demonstrated the feasibility of deriving these personality traits from digital footprints \cite{LiangZhu2017}, which provides a data-driven foundation for initializing heterogeneous agent populations.
Existing approaches often rely on synthetic distributions, lacking the empirical fidelity required for realistic simulation. Consequently, there is a critical need for grounding heterogeneous agent populations in empirical data.
\begin{figure}[t!]
\centering
\includegraphics[width=0.95\columnwidth]{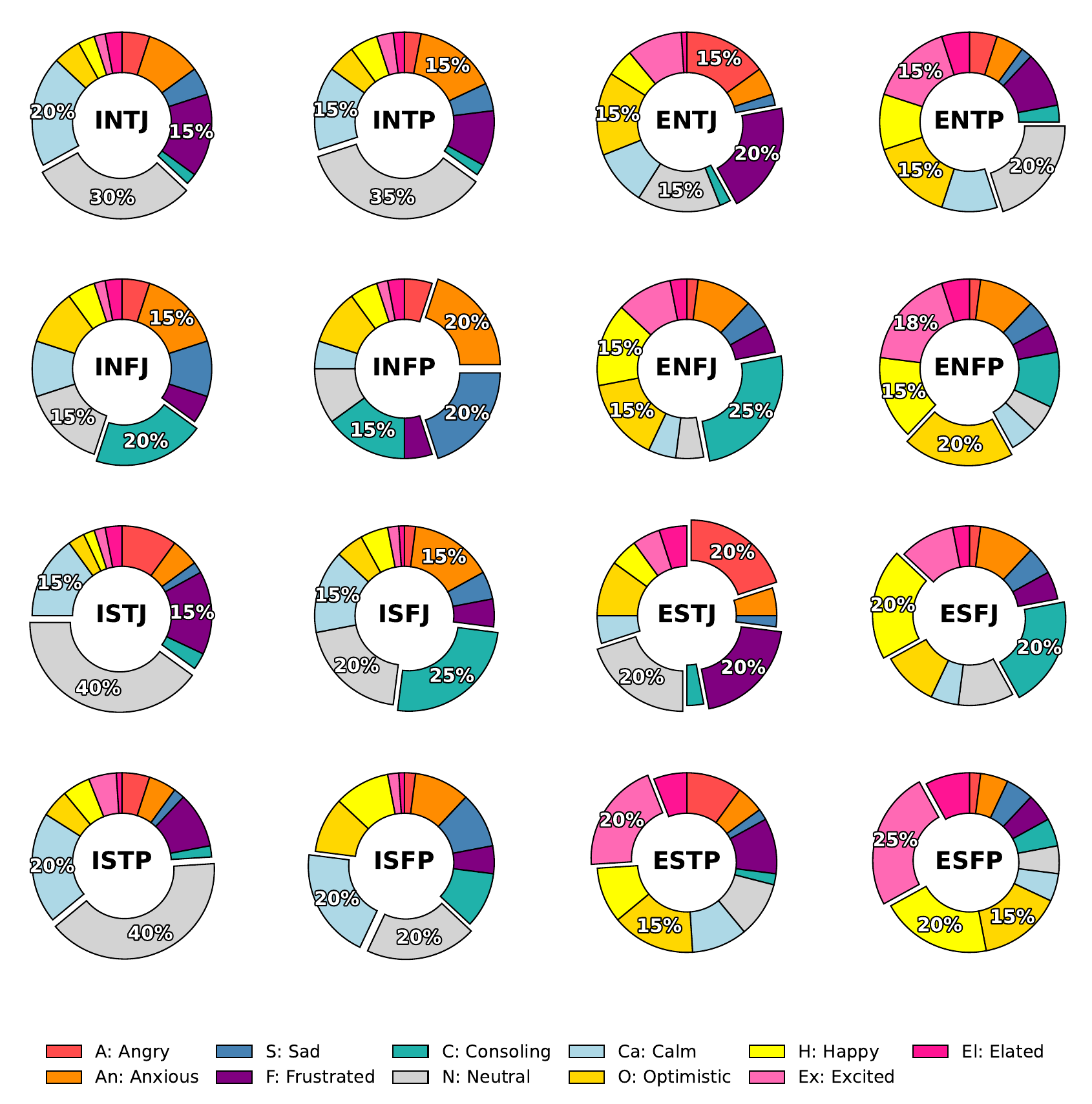}
\caption{Empirical Posterior Distributions of Affective States.}
\label{fig:emotion_piecharts}
\vspace{-1.5em}
\end{figure}

\begin{figure*}[t!]
\centering
\includegraphics[width=0.95\textwidth]{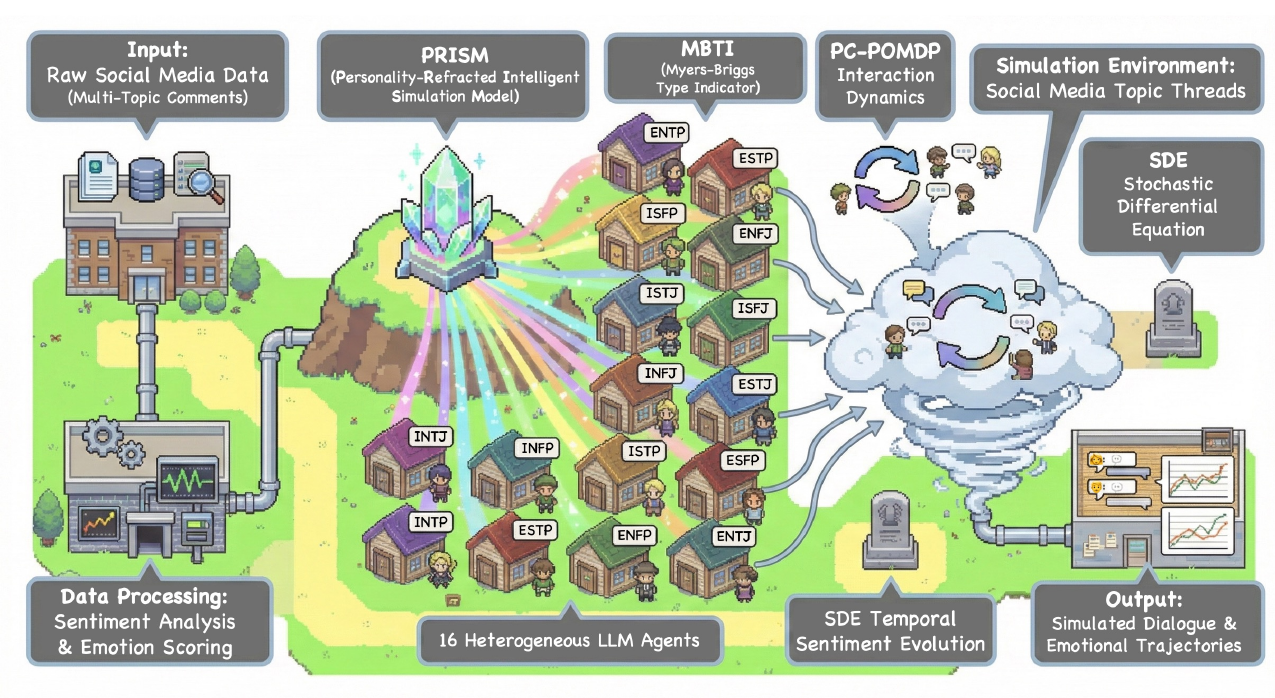}
\caption{The overall framework of PRISM.}
\label{fig:profiles}
\vspace{-1em}
\end{figure*}
To this end, we introduce the Personality-Refracted Intelligent Simulation Model (PRISM), a novel multi-agent framework that integrates multimodal large language models (MLLMs) with a mathematically grounded hybrid dynamical system. Unlike previous approaches that treat sentiment as a static property \cite{MadhoushiHamdanothers2015}, PRISM models the interaction loop as a coupling between continuous emotional evolution and discrete decision-making. Specifically, we formalize the micro-level affect using Stochastic Differential Equations (SDE) to capture the temporal decay and volatility of emotions \cite{wood_2018}. Simultaneously, we employ a Personality-Conditional Partially Observable Markov Decision Process (PC-POMDP) to govern the discrete logic of posting and replying \cite{pomdp2023}.

Our approach is grounded in a large-scale public dataset \cite{li2025rhythmopinionhawkesgraphframework} comprising 47,207 discussion topics and over 300,000 comments. By leveraging GPT-4o for automated personality and sentiment annotation, we construct statistically rigorous prior distributions (Fig. \ref{fig:emotion_piecharts}) that dictate how different agents refract incoming social signals. For instance, agents with Judging (J) traits exhibit higher stiffness in their emotional SDE. This leads to more stable opinion trajectories compared to spontaneous Perceiving (P) types. Three main contributions distinguish our work from prior research:
\begin{enumerate}
    \item We construct the first multi-agent system grounded in real-world multimodal commentary via MBTI profiling. Validated on over 47,000 topics, this framework effectively bridges the gap between theoretical psychology and realistic agent behavior.
    \item We propose a novel framework coupling SDE for continuous emotional evolution with PC-POMDP for discrete decision-making. This mechanism effectively resolves the state amnesia problem in traditional ABMs, enabling agents to maintain long-term behavioral coherence.
    \item Through controlled experiments, we demonstrate that PRISM achieves superior alignment with empirical human trajectories, reducing the distributional divergence by 66.7\% compared to homogeneous baselines and replicating emergent phenomena such as echo chamber formation with high fidelity.
\end{enumerate}

\section{Related Work}\label{related-work}

\subsection{Personality Modeling in Multi-Agent Systems}
The integration of personality models into multi-agent systems has evolved significantly from early rule-based approaches to contemporary data-driven methods. Initial attempts focused on implementing simplified personality dimensions, often reducing complex traits to single parameters affecting decision thresholds \cite{Ahrndt2019}. Recent advances have demonstrated the value of comprehensive personality frameworks, with the Big Five model dominating computational personality research \cite{CelliKarteljoreviSuhartonoothers2025}. However, MBTI's categorical structure offers distinct advantages for social media simulation, particularly in modeling discrete interaction patterns and communication styles. The work on deriving MBTI personalities in multi-agent systems \cite{Date2022} established foundational techniques for mapping type indicators to observable behaviors, though their approach lacked the data-driven validation we incorporate.

\subsection{Social Media Simulation Architectures}
Contemporary social media simulations increasingly recognize the importance of heterogeneous agent populations. The Y Social framework \cite{RossettiStellaCazabetAbramskiothers2024} demonstrated how MLLMs could power agent behaviors, though their personality implementation remained superficial. Network structure studies \cite{BakshyRosennMarlowAdamic2012,ICME2025} have traditionally dominated the field, often treating individual differences as noise rather than systematic variation. Our framework bridges these perspectives by combining network-aware interaction patterns with deep personality modeling. The emerging literature on MLLM-driven multi-agent simulations \cite{LiXuZhangMalthouse2024} informs our linguistic style modeling, though we extend these approaches with type-specific parameterization.

\subsection{MBTI Applications in Computational Social Science}
While MBTI has faced criticism in clinical psychology, its computational applications have shown promise in predicting online behavior patterns. Research on MBTI-based interaction networks \cite{AyyoubzadehShahnazariFazliothers2025} revealed systematic differences in connectivity preferences between types, directly informing our interaction probability calculations. Personality studies \cite{Zhang2024} provided crucial insights into how different personalities curate their online presence, which we operationalize through distinct linguistic style models. The synthesis of these findings addresses a critical gap in social simulation literature, where personality has often been treated as peripheral to structural factors.

\section{Methodology}
\label{sec:methodology}
\begin{algorithm}
\caption{PRISM Simulation Loop}
\label{alg:prism_master}
\begin{algorithmic}[1]
\REQUIRE Agent set $\mathcal{A}$, Social Graph $G=(\mathcal{V}, \mathcal{E})$, Time Horizon $T$, Step Size $\Delta t$, Thresholds $\Gamma_{\mathcal{T}}$, SDE Matrices $\{\boldsymbol{\Theta}, \boldsymbol{\Sigma}, \boldsymbol{\Psi}\}$, Dirichlet Prior $\alpha$, Observation Noise $\sigma_{obs}^2$
\ENSURE Emotional States $\mathbf{E} \in \mathbb{R}^{|\mathcal{A}| \times T \times d}$, History Logs $\mathcal{H}$

\textbf{Initialization:}
\FOR{each agent $i \in \mathcal{A}$}
    \STATE $\mathbf{e}_i^{(0)} \sim \text{Dirichlet}(\alpha)$
    \STATE $\mathcal{O}_i^{(0)} = (\mu_{i}^{(0)}, \tau_{i}^{(0)})$ 
    \STATE $H_i \leftarrow \emptyset$
\ENDFOR

\textbf{Main Simulation Loop:}
\FOR{$t = 0$ to $T-\Delta t$ step $\Delta t$}
    
    \FOR{each agent $i \in \mathcal{A}$}
      \STATE $d\mathbf{W}_t \sim \mathcal{N}(\mathbf{0}, \Delta t \mathbf{I})$
        \STATE $\mathbf{d}_{\text{drift}} \leftarrow \boldsymbol{\Theta}_{\mathcal{T}_i}(\boldsymbol{\mu}_{\mathcal{T}_i} - \mathbf{e}_i^{(t)}) \Delta t$
        \STATE $\tilde{\mathbf{e}}_i \leftarrow \mathbf{e}_i^{(t)} + \mathbf{d}_{\text{drift}} + \boldsymbol{\Sigma}_{\mathcal{T}_i} d\mathbf{W}_t$ 
    \ENDFOR

    \STATE $\mathcal{A}_{\text{active}} \leftarrow \emptyset$
    \FOR{each agent $i \in \mathcal{A}$}
        \STATE $U_i^{(t)} \leftarrow w_1 \sum_{j \in \mathcal{N}_i} \mathbb{I}(j \in \mathcal{A}_{\text{prev}}) + w_2 \|\mathbf{e}_i^{(t)} - \boldsymbol{\mu}_{\mathcal{T}_i}\|_2$
        \IF{$U_i^{(t)} > \Gamma_{\mathcal{T}_i}$}
            \STATE $\mathcal{A}_{\text{active}} \leftarrow \mathcal{A}_{\text{active}} \cup \{i\}$
        \ENDIF
    \ENDFOR

    \FOR{each speaker $s \in \mathcal{A}_{\text{active}}$}
        \STATE Context $C \leftarrow H_s \oplus \phi(\mathbf{e}_s^{(t)}) \oplus \rho(\mathcal{T}_s)$
        \STATE Message $\omega \sim \pi_{\text{MLLM}}(\cdot | C)$
        \STATE $H_s \leftarrow H_s \oplus \omega$
        
        \FOR{each listener $j \in \mathcal{N}_s$}
            \STATE $\mathbf{v}_{\text{msg}}, y_{\text{msg}} \leftarrow \boldsymbol{\Phi}(\omega)$
            
            \STATE $\tilde{\mathbf{e}}_j \leftarrow \tilde{\mathbf{e}}_j + \boldsymbol{\Psi}_{\mathcal{T}_j} \cdot \mathbf{v}_{msg}$
            
            \STATE $\tau_j^{(t+\Delta t)} \leftarrow \tau_j^{(t)} + \sigma_{obs}^{-2}$
            \STATE $\mu_j^{(t+\Delta t)} \leftarrow \frac{\tau_j^{(t)} \mu_j^{(t)} + \sigma_{obs}^{-2} y_{msg}}{\tau_j^{(t+\Delta t)}}$
            
            \STATE $H_j \leftarrow H_j \oplus \omega$
        \ENDFOR
    \ENDFOR
    
    \FOR{each agent $i \in \mathcal{A}$}
        \STATE $\mathbf{e}_i^{(t+\Delta t)} \leftarrow \Pi_{\mathcal{S}}(\tilde{\mathbf{e}}_i)$
    \ENDFOR
\ENDFOR
\end{algorithmic}
\end{algorithm}
\subsection{Data-Driven Emotional Parameterization}
\label{subsec:emotion_priors}


    
    
To ensure statistically grounded emotional responses, we initialized state transition matrices using empirical posterior distributions (Fig. \ref{fig:emotion_piecharts}). These priors were derived by aggregating dataset sentiment labels segmented via distant supervision \cite{gjurkovic2018reddit}. The distributions align with Jungian cognitive functions \cite{jung1921psychological} and five-factor model correlations \cite{mccrae_john_1992}. These prior distributions $P(s|\mathcal{T}_i)$ serve as bias terms, ensuring agents' emotional ``center of gravity'' remains true to their MBTI archetype:

\begin{itemize}
    \item \textbf{Rational Suppression in Thinking (T) Types:} Agents like ISTP and INTJ display a dominant \textit{Neutral} state ($30\%-40\%$). This reflects the ``tough-mindedness'' of the Thinking function \cite{Myers1976}, correlating with high emotional stability and a preference for objective communication.
    
    \item \textbf{Affective Resonance in Feeling (F) Types:} Conversely, Feeling types show expanded probability mass for \textit{Consoling} and \textit{Happy} states (e.g., ENFJ at 25\% Consoling). This empirically validates the ``Diplomat'' role theory \cite{keirsey1998please}, where the Feeling function drives agents towards social harmonization.
    
    \item \textbf{Assertiveness in Extraverted-Thinking (ET) Types:} ESTJ and ENTJ profiles show elevated levels of \textit{Frustration} and \textit{Anger}. This captures the assertive nature of the $Te$ function \cite{jung1921psychological}, which prioritizes efficiency over interpersonal harmony, often resulting in confrontational signals.
\end{itemize}

\subsection{Mathematical Formalization of Affective Priors}
\label{subsec:math_priors}

To rigorously quantify the personality-driven emotional divergence, we formalize the agent's baseline affect as a categorical distribution over the emotion set $\mathcal{E} = \{e_1, ..., e_K\}$.

\noindent \textbf{Bayesian Estimation of Emotion Distributions.}
The distributions presented in Fig. \ref{fig:emotion_piecharts} represent the \textit{empirical posterior probability} $P(e_k | \mathcal{T}_i)$. To ensure statistical robustness against sparse data in rare emotion categories, we employ dirichlet smoothing \cite{Blei2003} during parameter estimation:
\begin{equation}
\label{eq:dirichlet_smoothing}
\hat{\theta}_{i,k} = P(e_k | \mathcal{T}_i) = \frac{N(e_k, \mathcal{T}_i) + \alpha}{\sum_{j=1}^{K} N(e_j, \mathcal{T}_i) + K\alpha}
\end{equation}
where $N(e_k, \mathcal{T}_i)$ denotes the frequency count of emotion $k$ for type $i$, and $\alpha=1.0$ is the smoothing parameter. This formulation prevents overfitting to stochastic noise in the training data.

\noindent \textbf{Entropic Analysis of Rational Suppression.}
The fundamental difference between Thinking (T) and Feeling (F) agents is quantitatively validated by the Shannon entropy \cite{Shannon} of their respective distributions. We define the \textit{Affective Entropy} $H(\mathcal{T}_i)$ as:
\begin{equation}
\label{eq:affective_entropy}
H(\mathcal{T}_i) = - \sum_{k=1}^{K} \hat{\theta}_{i,k} \log_2 \hat{\theta}_{i,k}
\end{equation}
Thinking types exhibit minimized $H(\mathcal{T}_{Thinking})$ due to mass concentration on the \textit{Neutral} state, reflecting a predictable, low-variance affective policy. Conversely, Feeling types exhibit maximized $H(\mathcal{T}_{Feeling})$, indicating a higher degree of emotional complexity required for social bridging.


\noindent \textbf{Stochastic Quantization Regularization.} To mitigate the spurious precision of continuous priors exceeding the effective resolution of unsupervised sentiment analysis, we employ probability coarse-graining~\cite{domenico_2025}. Affective parameters are projected onto a discretized manifold with step $\epsilon=0.05$, where the regularized $\hat{\theta}_{i,k}^{final}$ is computed as:
\begin{equation}
\hat{\theta}_{i,k}^{final} = \frac{1}{Z} \cdot \epsilon \left\lfloor \frac{\hat{\theta}_{i,k}}{\epsilon} \right\rceil
\end{equation}
where $\lfloor \cdot \rceil$ denotes the nearest integer function, and $Z$ serves as the normalization factor to ensure the probability distribution constraints are maintained after quantization.

\subsection{Hybrid Interaction Dynamics}
\label{subsec:hybrid_dynamics}

To capture the complex interplay between internal affect and external communication, we formalize the agent interaction as a jump-diffusion process \cite{arias2023newselfexcitingjumpdiffusionprocess} coupled with a discrete control policy.

\noindent \textbf{Continuous Emotional Evolution (SDE Layer).} Let $\mathbf{e}_i^{(t)} \in \mathbb{R}^d$ denote the continuous emotional state vector of agent $i$. We model its temporal evolution using a multivariate ornstein-uhlenbeck process \cite{oksendal2003stochastic}, which captures the psychological tendency of mean-reversion subject to cognitive noise:
\begin{equation}
d\mathbf{e}_i^{(t)} = \mathbf{\Theta}_{\mathcal{T}_i} (\boldsymbol{\mu}_{\mathcal{T}_i} - \mathbf{e}_i^{(t)}) dt + \mathbf{\Sigma}_{\mathcal{T}_i} d\mathbf{W}_t + \mathbf{J}_{in}(t)
\end{equation}
where $\boldsymbol{\mu}_{\mathcal{T}_i}$ is the baseline centroid, $\mathbf{\Theta}_{\mathcal{T}_i}$ is the stiffness matrix. $\mathbf{\Sigma}_{\mathcal{T}_i}$ represents intrinsic volatility and $d\mathbf{W}_t$ is a Wiener process. The state trajectory between interaction events $t_k$ and $t_{k+1}$ is analytically given by the It\^o integral. To fit the column width, we express it as:
\begin{equation}
\begin{split}
\mathbf{e}_i^{(t)} &= \boldsymbol{\mu}_{\mathcal{T}_i} + e^{-\mathbf{\Theta}_{\mathcal{T}_i}(t-t_k)} (\mathbf{e}_i(t_k) - \boldsymbol{\mu}_{\mathcal{T}_i}) \\
&\quad + \int_{t_k}^{t} e^{-\mathbf{\Theta}_{\mathcal{T}_i}(t-u)} \mathbf{\Sigma}_{\mathcal{T}_i} d\mathbf{W}_u
\end{split}
\end{equation}
\noindent \textbf{Discrete Interaction Logic (PC-POMDP Layer).} Social interactions act as discrete perturbations. The PC-POMDP \cite{pomdp2023} tuple $\langle \mathcal{S}, \Omega, \pi_{\mathcal{T}} \rangle$ governs the discrete control logic.

\noindent \textbf{Interaction Impulse (Coupling Term).}
When agent $i$ receives a message $\omega$ at time $t$, it triggers the jump term $\mathbf{J}_{in}(t)$. Let $\mathbf{v}(\omega)$ be the valence-arousal vector. The impulse is modeled using the dirac delta function \cite{Zhang_2020}:
\begin{equation}
\mathbf{J}_{in}(t) = \sum_{k} \delta(t - t_k) \cdot \mathbf{\Psi}_{\mathcal{T}_i} \cdot \mathbf{v}(\omega_k)
\end{equation}
Where $\mathbf{\Psi}_{\mathcal{T}_i}$ is the susceptibility matrix, in which Feeling (F) types exhibit higher spectral norms ($||\mathbf{\Psi}_{F}|| > ||\mathbf{\Psi}_{T}||$), quantifying their higher sensitivity to emotional contagion.

\noindent \textbf{Policy Execution (Response Generation).}
The agent generates a response $x_t$ via a MLLM, conditioned on both the dialogue history $H_t$ and the continuous emotional state $\mathbf{e}_i^{(t)}$:
\begin{equation}
\begin{gathered}
\mathbf{z}_t = \mathcal{E}(H_t) \oplus \phi(\mathbf{e}_i^{(t)}) \oplus \rho(\mathcal{T}_i) \\
P(x_t \mid H_t, \mathbf{e}_i^{(t)}) = \operatorname{softmax}( \mathbf{W}_{o} \mathcal{F}_{\theta}(\mathbf{z}_t) )
\end{gathered}
\end{equation}
where $\phi(\cdot)$ projects the continuous emotional vector into the MLLM's latent space, and $\rho(\mathcal{T}_i)$ enforces MBTI-specific linguistic constraints.

\noindent \textbf{Response Activation Mechanism.} Unlike round-robin mechanisms, PRISM employs a threshold-based activation model to determine interaction sequences. An agent $i$ will generate a reply if the interaction impulse $U_{ij}(t)$ exceeds the specific activation threshold $\Gamma_{\mathcal{T}_i}$:

\begin{equation}
\label{eq:activation_condition}
\text{Action}_i^t = 
\begin{cases} 
\text{Reply}, & \text{if } U_{ij}(t) > \Gamma_{\mathcal{T}_i} \\
\text{Silence}, & \text{otherwise}
\end{cases}
\end{equation}

The Impulse Function $U_{ij}(t)$ combines structural social pressure with the dynamic emotional energy:
\begin{equation}
U_{ij}(t) = w_1 G_{ij} + w_2 \|\mathbf{e}_i^{(t)} - \boldsymbol{\mu}_{\mathcal{T}_i}\|_2 + w_3 \mathbb{I}_{mention}
\end{equation}
The term $\|\mathbf{e}_i^{(t)} - \boldsymbol{\mu}_{\mathcal{T}_i}\|_2$ creates a realistic feedback loop: agents in agitated states are mathematically more likely to overcome their threshold $\Gamma_{\mathcal{T}_i}$ and engage in debate.

\begin{table}[t!]
    \centering
    \caption{Performance comparison on EQ-Bench.}
    \label{tab:eq_bench_results}
    \begin{tabular}{l c c c}
    \toprule
    \textbf{Model / Agent} & \textbf{EQ-Bench Score} & \textbf{$\Delta$ vs Base} & \textbf{Attribute Group} \\
    \midrule
    \textit{GPT-5.1} & \textit{88.14} & \textit{+18.02} & \textit{Topline SOTA} \\
    \textit{Qwen3-8B-VL} & \textit{70.12} & \textit{-} & \textit{Base Model} \\
    \midrule
    \rowcolor{gray!10}
    \multicolumn{4}{l}{\textit{Feeling (F) Agents (Empathy)}} \\
    ENFJ & 81.48 & +11.36 & Diplomats \\
    ENFP & 80.78 & +10.66 & Diplomats \\
    INFJ & 80.36 & +10.24 & Diplomats \\
    ESFJ & 79.68 & +9.56 & Sentinels \\
    ESFP & 79.02 & +8.90 & Explorers \\
    INFP & 78.46 & +8.34 & Diplomats \\
    ISFJ & 77.83 & +7.71 & Sentinels \\
    ISFP & 77.43 & +7.31 & Explorers \\
    \midrule
    \rowcolor{gray!10}
    \multicolumn{4}{l}{\textit{Thinking (T) Agents (Rational)}} \\
    ENTJ & 69.06 & -1.06 & Analysts \\
    ENTP & 68.18 & -1.94 & Analysts \\
    INTJ & 67.42 & -2.70 & Analysts \\
    INTP & 66.29 & -3.83 & Analysts \\
    ESTJ & 65.09 & -5.03 & Sentinels \\
    ESTP & 64.28 & -5.84 & Explorers \\
    ISTJ & 63.11 & -7.01 & Sentinels \\
    ISTP & 61.97 & -8.15 & Explorers \\
    \bottomrule
    \end{tabular}
    \vspace{-1em}
\end{table}
\section{Experiments}
\subsection{Experimental Setup}
\label{subsec:setup}

\noindent \textbf{Dataset.} Our simulation used a large-scale public dataset \cite{li2025rhythmopinionhawkesgraphframework} that includes 47,207 discussion topics, along with user comments and corresponding sentiment scores. To establish statistically rigorous priors, we employed GPT-4o \cite{gpt4osystemcard} to annotate the corpus with MBTI labels. 

\noindent \textbf{Implementation Details.} Experiments were conducted on a cluster with 8$\times$ NVIDIA A100 (80GB) GPUs. To guarantee robust role adherence, we fine-tuned the Qwen3-8B-VL foundational model \cite{qwen3technicalreport} for 5 epochs using Role-Conditioned Instruction Tuning (RoleCIT) \cite{wang-etal-2024-rolellm} to construct the 16 personality-specific agent profiles.

\noindent \textbf{Evaluation Metrics.} We evaluate framework performance using Polarity Error (MAE) to measure the distributional divergence from ground-truth sentiment trajectories, and Consistency ($\rho$) to quantify the Spearman rank correlation (\ref{tab:method_comparison}) between generated agent behaviors and theoretical personality priors.

\noindent \textbf{Baseline Methods.} We benchmark PRISM against three established frameworks: Homogeneous Rational Agents (HRA) \cite{MuellerFrank2013} as a control group with deterministic rules, the Random Behavior Model (RBM) \cite{Wasserman1980} utilizing stochastic probabilities, and the Big Five Personality Model (BFPM) \cite{mccrae_john_1992}, a continuous trait-based model representing the current state-of-the-art in computational personality simulation.

\begin{table}[t!]
\centering
\caption{Comparative performance on sentiment polarity.}
\label{tab:method_comparison}
\resizebox{\linewidth}{!}{%
\begin{tabular}{l|c|c|c}
\toprule
\textbf{Method} & \textbf{Polarity Error} (MAE) $\downarrow$ & \textbf{Consistency $\rho$} $\uparrow$ & $p$-value \\
\midrule
\textbf{HRA} \cite{MuellerFrank2013} & 0.42 $\pm$ 0.08 & 0.154 & 0.12 \\
\textbf{RBM} \cite{Wasserman1980} & 0.55 $\pm$ 0.11 & 0.021 & 0.85 \\
\textbf{BFPM} \cite{mccrae_john_1992} & 0.21 $\pm$ 0.05 & 0.613 &  0.05 \\
\midrule
\rowcolor{gray!10}
\textbf{PRISM (Ours)} & \textbf{0.14 $\pm$ 0.03} & \textbf{0.782} & $\mathbf{< 0.01}$ \\
\bottomrule
\end{tabular}%
}
\vspace{-1em}
\end{table}
\subsection{Agent Capability Assessment}
\label{subsec:eq_bench}

\begin{figure}
\centering
\includegraphics[width=0.95\columnwidth]{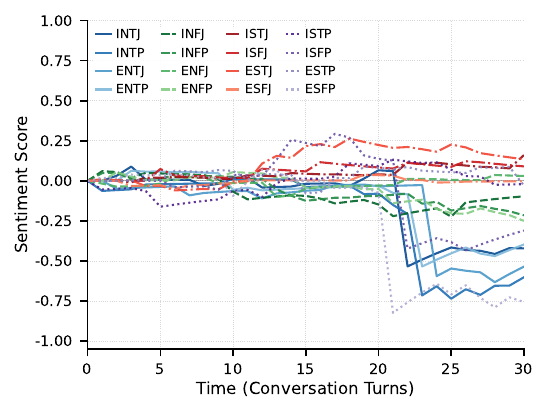}
\caption{Change in average sentiment score of all agents over time during simulation of discussions on a controversial topic.}
\label{fig:sentiment_evolution}
\vspace{-1em}
\end{figure}
Before analyzing system-level dynamics, we assessed individual agent capabilities using EQ-Bench \cite{eqbenchmark}, benchmarking our 16 MBTI-specific agents against the Qwen3-8B-VL base model and GPT-5.1 \cite{openai2025gpt51}. As shown in Table \ref{tab:eq_bench_results}, the results reveal a functional divergence necessary for realistic social roles: agents with Feeling traits achieved significant gains in emotional understanding. For instance, the ENFJ agent achieved an EQ-Bench score of 81.48, marking a 11.36 point improvement over the base model (70.12), confirming that personality-driven prompting unlocks latent empathy, while Thinking types exhibited a rational penalty, such as ISTP (61.97) and ISTJ (63.11), scored lower than the base model, validating their fidelity to objective, low-emotion profiles.
\begin{table}[t]
\centering
\caption{Ablation study on personality dimensions.}
\label{tab:ablation}
\resizebox{\linewidth}{!}{%
\begin{tabular}{lcccc}
\toprule
\textbf{Model Variant} & \multicolumn{2}{c}{\textbf{Polarity Error (MAE)} $\downarrow$} & \multicolumn{2}{c}{\textbf{Consistency $\rho$} $\uparrow$} \\
\cmidrule(lr){2-3} \cmidrule(lr){4-5}
& \textbf{Score} & \textbf{$\Delta$ (\%)} & \textbf{Score} & \textbf{$\Delta$ (\%)} \\
\midrule
\rowcolor{gray!10}
\textbf{PRISM (Full Model)} & \textbf{0.140} & -- & \textbf{0.782} & -- \\
\midrule
\quad w/o Sensing (S/N) & 0.160 & \textcolor{gray}{(+14.3\%)} & 0.741 & \textcolor{gray}{(-5.2\%)} \\
\quad w/o Judging (J/P) & 0.190 & \textcolor{gray}{(+35.7\%)} & 0.692 & \textcolor{gray}{(-11.5\%)} \\
\quad w/o Extraversion (E/I) & 0.220 & (+57.1\%) & 0.588 & (-24.8\%) $^{\ast}$ \\
\quad w/o Thinking (T/F) & 0.320 & \textbf{(+128.6\%)} & 0.453 & \textbf{(-42.1\%)} $^{\ast\ast}$ \\
\bottomrule
\end{tabular}%
}
\vspace{-1em}
\end{table}

\subsection{Quantitative System Evaluation}
\label{subsec:quantitative}
\subsubsection{Macro-Level Quantitative Analysis}
We evaluated the collective behavior of the system across 1,000 unseen controversial topics. As detailed in Table \ref{tab:method_comparison}, PRISM significantly outperforms baselines, achieving a Polarity Error of 0.14 (vs. BFPM 0.21, HRA 0.42) while maintaining high behavioral stability ($\rho=0.782, p<0.01$). This empirical evidence confirms that discrete MBTI policies provide superior behavioral regularization compared to continuous personality traits, effectively mitigating stochastic drift in long-horizon simulations.

\subsubsection{Qualitative Case Study}
\begin{figure}[t]
\centering
\includegraphics[width=0.95\columnwidth]{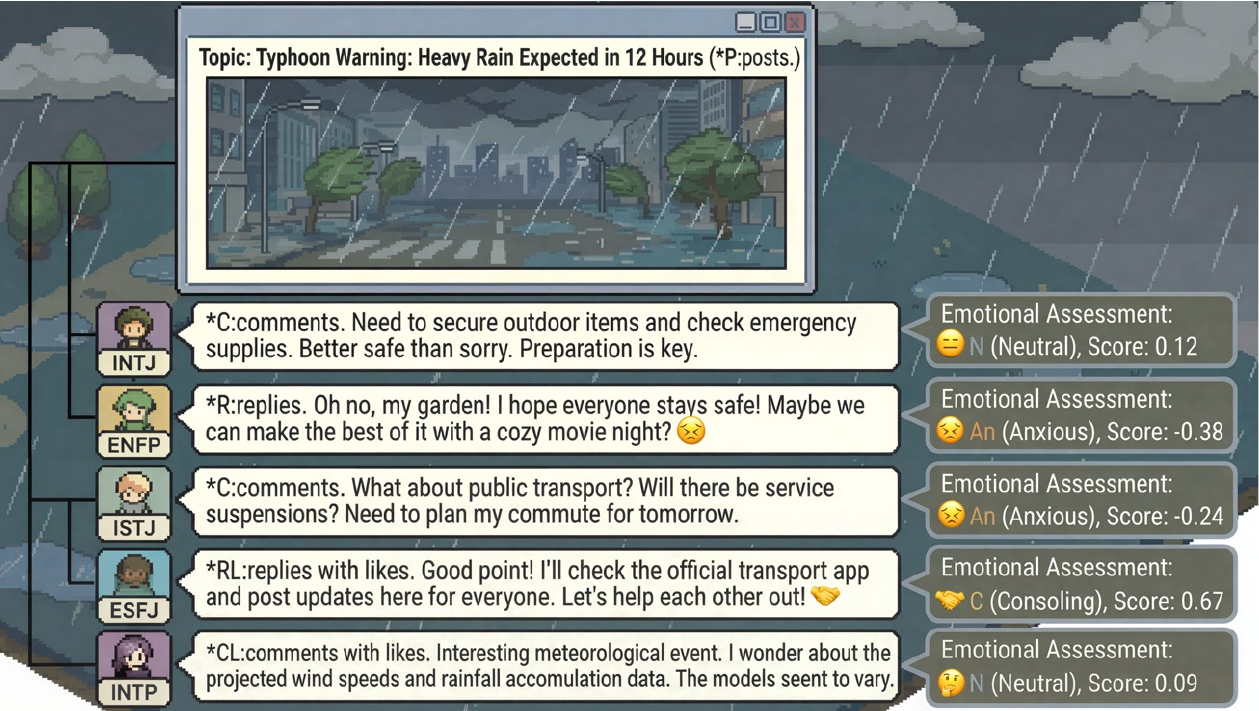}
\caption{Case study visualization. A multi-turn dialogue simulation of a thread.}
\label{fig:dialogue}
\vspace{-1em}
\end{figure}

To visualize micro-level interaction mechanics, Fig. \ref{fig:sentiment_evolution} tracks the real-time sentiment trajectories of 16 agents during a \textit{``Citywide Heavy Rain Warning''} scenario (the dialogue is shown in Fig. \ref{fig:dialogue}). The simulation reveals a critical phase transition at $t=21$:

\begin{itemize}
    \item \textbf{Trigger Event:} An ENTP agent initiates a negative contagion wave by questioning infrastructure capacity (\textit{``...downtown pumps will fail within 2 hours''}). This utterance generates a high-valence impulse $\mathbf{J}_{in}(t)$ within the SDE model.
    \item \textbf{Divergent Responses:} This impulse triggers a bifurcation. Analyst (NT) types (Blue/Purple lines) experience a sharp sentiment drop to $-0.75$, reflecting logical anxiety. Conversely, Diplomat (NF) types (Red/Green lines) demonstrate emotional resilience (sentiment $\approx 0.1$), actively buffering the panic with constructive support.
\end{itemize}


\subsection{Ablation Study}
\label{subsec:ablation}

To disentangle the contribution of specific personality dimensions to collective emotional dynamics, we conducted a systematic ablation study. We defined the full model (all MBTI dimensions active) as the baseline and neutralized each dimension individually to observe the degradation in system fidelity. Table \ref{tab:ablation} reports the impact on both polarity error and behavioral consistency ($\rho$). We employ the Wilcoxon signed-rank test \cite{wtest} to assess the statistical significance of the performance deviation from the full model trajectories. The results reveal distinct roles for different personality traits in shaping social discourse:

\begin{itemize}
\item \textbf{Polarity Determinants (T/F):} This dimension governs sentiment fidelity. Its ablation triggers a maximum 128.6\% surge in polarity error (Table~\ref{tab:ablation}), validating that the interplay between rational suppression (T) and affective resonance (F) dictates opinion trajectories.

\item \textbf{Interaction Catalysts (E/I):} The E/I dimension regulates temporal dynamics, with its removal causing a significant consistency degradation ($p < 0.01$). This confirms that E-types serve as pivotal emotional hubs that accelerate information diffusion and drive conversation rhythm.
\end{itemize}

\section{Conclusion}
\label{sec:conclusion}
We introduced PRISM, a hybrid framework synergistically coupling MLLMs with stochastic dynamical systems (SDE and PC-POMDP) to simulate psychologically heterogeneous social discourse. We demonstrate that discrete MBTI priors provide superior behavioral regularization compared to continuous traits, catalyzing the emergence of distinct cognitive roles. Empirically, PRISM achieves significant alignment with human ground truth ($\rho=0.782$, $p$-value$<$0.01) and a 66.7\% reduction in polarity error, faithfully replicating complex emergent phenomena such as rational suppression and affective resonance. These findings establish PRISM as a robust proxy for decoding the mechanistic underpinnings of online polarization.


\bibliographystyle{IEEEbib}
\bibliography{icme2026references}

@book{UhrmacherWeyns2018,
  author = {AM Uhrmacher and D Weyns},
  title = {Multi-Agent systems: Simulation and applications},
  publisher = {CRC Press},
  year = {2018}
}

@article{Chen2019,
author = {Zhuo Chen},
title = {An Agent-Based Model for Information Diffusion over Online Social Networks},
journal = {Papers in Applied Geography},
volume = {5},
number = {1-2},
pages = {77--97},
year = {2019},
publisher = {Routledge},
doi = {10.1080/23754931.2019.1619193},
}

@book{Myers1976,
title={The Myers-Briggs Type Indicator: Manual}, publisher ={Consulting Psychologists Press.}, author={Myers, Isabel Briggs}, year={1962} }

@article{mccrae_john_1992, title={An Introduction to the five-factor Model and Its Applications}, volume={60}, DOI={https://doi.org/10.1111/j.1467-6494.1992.tb00970.x}, number={2}, journal={Journal of Personality}, author={McCrae, Robert R. and John, Oliver P.}, year={1992}, pages={175–215} }

@book{boyle_saklofske_matthews_2015, title={Measures of personality and social psychological constructs}, ISBN={9780123869159}, publisher={Academic Press, An Imprint Of Elsevier}, author={Boyle, Gregory John and Saklofske, Donald H and Matthews, Gerald}, year={2015}, pages={752–776} }

@book{boer_2013, address={Witney}, title={Whole Brain Learning in Higher Education : Evidence-Based Practice.}, ISBN={9781843347422}, publisher={Woodhead Publishing Ltd}, author={Ann-Louise De Boer}, year={2013} }

@book{LiangZhu2017,
author = {Liang, Hai and Zhu, Jonathan J. H.},
publisher = {John Wiley \& Sons, Ltd},
isbn = {9781118901731},
title = {Big Data, Collection of (Social Media, Harvesting)},
booktitle = {The International Encyclopedia of Communication Research Methods},
chapter = {},
pages = {1-18},
doi = {https://doi.org/10.1002/9781118901731.iecrm0015},
year = {2017}
}

@inproceedings{MadhoushiHamdanothers2015,
  author={Madhoushi, Zohreh and Hamdan, Abdul Razak and Zainudin, Suhaila},
  booktitle={2015 Science and Information Conference (SAI)}, 
  title={Sentiment analysis techniques in recent works}, 
  year={2015},
  doi={10.1109/SAI.2015.7237157}}

@article{wood_2018, title={Modeling Intraindividual Dynamics Using Stochastic Differential Equations: Age Differences in Affect Regulation}, volume={73}, DOI={https://doi.org/10.1093/geronb/gbx013}, number={1}, journal={The journals of gerontology. Series B, Psychological sciences and social sciences.}, author={Wood, Julie and Oravecz, Zita and others}, year={2018}, pages={171–184} }

@article{pomdp2023,
  author={Lauri, Mikko and Hsu, David and Pajarinen, Joni},
  journal={IEEE Transactions on Robotics}, 
  title={Partially Observable Markov Decision Processes in Robotics: A Survey}, 
  year={2023},
  volume={39},
  number={1},
  pages={21-40},
  doi={10.1109/TRO.2022.3200138}}

@article{li2025rhythmopinionhawkesgraphframework,
      title={Rhythm of Opinion: A Hawkes-Graph Framework for Dynamic Propagation Analysis}, 
      author={Yulong Li and Zhixiang Lu and others},
      year={2025},
      eprint={2504.15072},
      archivePrefix={arXiv},
      primaryClass={cs.SI},
      journal={arXiv preprint arXiv:2504.15072}, 
}

@article{Ahrndt2019,
author = {Xue, Junxiao and others},
title = {A Personality-based Model of Emotional Contagion and Control in Crowd Queuing Simulations},
year = {2023},
publisher = {Association for Computing Machinery},
volume = {33},
number = {1–2},
issn = {1049-3301},
url = {https://doi.org/10.1145/3577589},
doi = {10.1145/3577589},
journal = {ACM Trans. Model. Comput. Simul.},
articleno = {6},
numpages = {23}
}

@inproceedings{CelliKarteljoreviSuhartonoothers2025,
    title = "On Text-based Personality Computing: Challenges and Future Directions",
    author = "Fang, Qixiang and others",
    booktitle = "Findings of the Association for Computational Linguistics: ACL",
    year = "2023",
    doi = "10.18653/v1/2023.findings-acl.691",
}

@inproceedings{Date2022,
author = {Braz, Luiz Fernando and others},
title = {Simulating Work Teams Using MBTI Agents},
year = {2022},
isbn = {978-3-031-22946-6},
publisher = {Springer-Verlag},
url = {https://doi.org/10.1007/978-3-031-22947-3_5},
doi = {10.1007/978-3-031-22947-3_5},
booktitle = {Multi-Agent-Based Simulation XXIII: 23rd International Workshop},
pages = {57–69},
numpages = {13}
}

@article{RossettiStellaCazabetAbramskiothers2024,
      title={Y Social: an LLM-powered Social Media Digital Twin}, 
      author={Giulio Rossetti and others},
      year={2024},
      eprint={2408.00818},
      archivePrefix={arXiv},
      primaryClass={cs.AI},
      journal={arXiv preprint arXiv:2408.00818}, 
}

@inproceedings{BakshyRosennMarlowAdamic2012,
author = {Bakshy, Eytan and Rosenn, Itamar and Marlow, Cameron and Adamic, Lada},
title = {The role of social networks in information diffusion},
year = {2012},
isbn = {9781450312295},
publisher = {Association for Computing Machinery},
doi = {10.1145/2187836.2187907},
booktitle = {Proceedings of the 21st International Conference on World Wide Web},
pages = {519–528},
numpages = {10},
series = {WWW '12}
}

@INPROCEEDINGS{ICME2025,
  author={Shen, Lingzhi and Long, Yunfei and Cai, Xiaohao and Chen, Guanming and Wang, Yuhan and Razzak, Imran and Jameel, Shoaib},
  booktitle={2025 IEEE International Conference on Multimedia and Expo (ICME)}, 
  title={LL4G: Self-Supervised Dynamic Optimization for Graph-Based Personality Detection}, 
  year={2025},
  doi={10.1109/ICME59968.2025.11209311}}

@article{LiXuZhangMalthouse2024,
      title={Large Language Model-driven Multi-Agent Simulation for News Diffusion Under Different Network Structures}, 
      author={Xinyi Li and Yu Xu and Yongfeng Zhang and Edward C. Malthouse},
      year={2024},
      eprint={2410.13909},
      archivePrefix={arXiv},
      primaryClass={cs.SI},
      journal={arXiv preprint arXiv:2410.13909}, 
}

@article{AyyoubzadehShahnazariFazliothers2025,
      title={Can Invisible Psychological Traits Organize Visible Network Structure? A Complex Network Analysis of Myers-Briggs Type Indicator-Based Interaction Patterns in Anonymous Social Networks}, 
      author={Seyed Moein Ayyoubzadeh and Kourosh Shahnazari and Mohammadamin Fazli and Mohammadali Keshtparvar},
      year={2025},
      eprint={2503.19704},
      archivePrefix={arXiv},
      primaryClass={cs.SI},
      journal={arXiv preprint arXiv:2503.19704}, 
}

@inproceedings{Zhang2024,
author = {Roy, Quentin and Ghafurian, Moojan and Li, Wei and Hoey, Jesse},
title = {Users, Tasks, and Conversational Agents: A Personality Study},
year = {2021},
isbn = {9781450386203},
doi = {10.1145/3472307.3484173},
booktitle = {Proceedings of the 9th International Conference on Human-Agent Interaction},
pages = {174–182},
numpages = {9},
series = {HAI '21}
}

@article{MuellerFrank2013,
  author = {M Mueller‐Frank},
  title = {A general framework for rational learning in social networks},
  journal = {Theoretical Economics},
  year = {2013}
}

@article{Wasserman1980,
  author = {S Wasserman},
  title = {Analyzing social networks as stochastic processes},
  journal = {Journal of the American statistical association},
  year = {1980}
}

@inproceedings{gjurkovic2018reddit,
    title = "{R}eddit: A Gold Mine for Personality Prediction",
    author = "Gjurkovi{\'c}, Matej  and
      {\v{S}}najder, Jan",
    booktitle = "Proceedings of the Second Workshop on Computational Modeling of People{'}s Opinions, Personality, and Emotions in Social Media",
    year = "2018",
    doi = "10.18653/v1/W18-1112",
    pages = "87--97",
}

@book{keirsey1998please, address={Del Mar, Ca}, title={Please understand me II : temperament, character, intelligence}, ISBN={9781885705020}, publisher={Prometheus Nemesis}, author={Keirsey, David}, year={1998} }

@article{jung1921psychological,
      title={How Jungian Cognitive Functions Explain MBTI Type Prevalence in Computer Industry Careers}, 
      author={Arya VarastehNezhad and Behnam Agahi and Soroush Elyasi and Reza Tavasoli and Hamed Farbeh},
      year={2025},
      eprint={2504.17248},
      archivePrefix={arXiv},
      primaryClass={cs.CY},
      journal={arXiv preprint arXiv:2504.17248}, 
}

@book{oksendal2003stochastic,
   title={Ornstein-Uhlenbeck process and generalizations: Particle dynamics under comb constraints and stochastic resetting},
   volume={107},
   ISSN={2470-0053},
   url={http://dx.doi.org/10.1103/PhysRevE.107.054129},
   DOI={10.1103/physreve.107.054129},
   journal={Physical Review E},
   publisher={American Physical Society (APS)},
   author={Trajanovski, Pece and Jolakoski, Petar and Zelenkovski, Kiril and Iomin, Alexander and Kocarev, Ljupco and Sandev, Trifce},
   year={2023} }

@article{wtest,
  author={García, Salvador and others},
  journal={IEEE Transactions on Knowledge and Data Engineering}, 
  title={A Survey of Discretization Techniques: Taxonomy and Empirical Analysis in Supervised Learning}, 
  year={2013},
  volume={25},
  number={4},
  pages={734-750},
  doi={10.1109/TKDE.2012.35}}

@inproceedings{Blei2003,
author = {Asuncion, Arthur and Welling, Max and Smyth, Padhraic and Teh, Yee Whye},
title = {On smoothing and inference for topic models},
year = {2009},
isbn = {9780974903958},
publisher = {AUAI Press},
booktitle = {Proceedings of the Twenty-Fifth Conference on Uncertainty in Artificial Intelligence},
pages = {27–34},
numpages = {8},
series = {UAI '09}
}

@article{Zhang_2020,
   title={Dirac Delta Function of Matrix Argument},
   volume={60},
   ISSN={1572-9575},
   url={http://dx.doi.org/10.1007/s10773-020-04598-8},
   DOI={10.1007/s10773-020-04598-8},
   number={7},
   journal={International Journal of Theoretical Physics},
   publisher={Springer Science and Business Media LLC},
   author={Zhang, Lin},
   year={2020}, pages={2445–2472} }

@article{Shannon,
      title={Shannon's entropy and its Generalizations towards Statistics, Reliability and Information Science during 1948-2018}, 
      author={Asok K Nanda and Shovan Chowdhury},
      year={2019},
      eprint={1901.09779},
      archivePrefix={arXiv},
      primaryClass={stat.OT},
      journal={arXiv preprint arXiv:1901.09779}, 
}

@article{domenico_2025, title={Coarse-graining network flow through statistical physics and machine learning}, volume={16}, url={https://www.nature.com/articles/s41467-025-56034-2#citeas}, DOI={https://doi.org/10.1038/s41467-025-56034-2}, number={1}, journal={Nature Communications}, publisher={Nature Portfolio}, author={Zhang, Zhang and Arsham Ghavasieh and Zhang, Jiang and Domenico, Manlio De}, year={2025}, pages={1605–1605} }

@article{arias2023newselfexcitingjumpdiffusionprocess,
      title={A new self-exciting jump-diffusion process for option pricing}, 
      author={Luis A. Souto Arias and Pasquale Cirillo and Cornelis W. Oosterlee},
      year={2023},
      eprint={2205.13321},
      archivePrefix={arXiv},
      primaryClass={q-fin.PR},
      journal={arXiv preprint arXiv:2205.13321}, 
}

@article{qwen3technicalreport,
      title={Qwen3 Technical Report}, 
      author={Qwen team},
      year={2025},
      eprint={2505.09388},
      archivePrefix={arXiv},
      primaryClass={cs.CL},
      journal={arXiv preprint arXiv:2505.09388}, 
}

@article{gpt4osystemcard,
      title={GPT-4o System Card}, 
      author={OpenAI},
      year={2024},
      eprint={2410.21276},
      archivePrefix={arXiv},
      primaryClass={cs.CL},
      journal={arXiv preprint arXiv:2410.21276}, 
}

@inproceedings{wang-etal-2024-rolellm,
    title = "{R}ole{LLM}: Benchmarking, Eliciting, and Enhancing Role-Playing Abilities of Large Language Models",
    author = "Wang and others",
    booktitle = "Findings of the Association for Computational Linguistics: ACL",
    year = "2024",
    doi = "10.18653/v1/2024.findings-acl.878",
    pages = "14743--14777",
}

@article{eqbenchmark,
      title={EQ-Bench: An Emotional Intelligence Benchmark for Large Language Models}, 
      author={Samuel J. Paech},
      year={2024},
      eprint={2312.06281},
      archivePrefix={arXiv},
      primaryClass={cs.CL},
      journal={arXiv preprint arXiv:2312.06281}, 
}

@techreport{openai2025gpt51,
  title       = {{GPT-5.1 Instant and GPT-5.1 Thinking System Card Addendum}},
  author      = {{OpenAI}},
  institution = {OpenAI},
  year        = {2025},
  url         = {https://openai.com/index/gpt-5-system-card-addendum-gpt-5-1/},
}


\end{document}